\documentclass{article}

\usepackage{microtype}
\usepackage{graphicx}
\usepackage{subfigure}
\usepackage{booktabs} 

\usepackage[nohyperref, accepted]{icml2019}

\icmltitlerunning{Tensor Monte Carlo: Particle Methods for the GPU Era}

\usepackage{tikz}
\usepackage{amsmath,amssymb}
\usetikzlibrary{calc,arrows.meta}
\usepackage{natbib}

\renewcommand{\vec}[1]{\mathbf{#1}}

\newcommand{\Q}{\mat{Q}}

\newcommand{\N}{\mathcal{N}\b}
\renewcommand{\P}[2][]{\text{P}_{#1}\!\b{#2}}
\renewcommand{\Q}[2][]{\text{Q}_{#1}\!\b{#2}}

\renewcommand{\sb}[1]{\left[ #1 \right]}

\renewcommand{\b}[1]{\left( #1 \right)}
\newcommand{\E}[1][]{\text{E}_{{#1}}\!\sb}

\newcommand{\Dkl}[1]{D_{\text{KL}}\b{#1}}

\newcommand{\s}{\sigma}

\newcommand{\f}{\vec{f}}

\newcommand{\z}{\vec{z}}
\renewcommand{\L}{\mathcal{L}}

\renewcommand{\d}[2][]{\frac{\partial #1}{\partial #2}}

\newcommand{\wb}{\bar{w}}
\newcommand{\wh}{\hat{w}}

\newcommand{\kv}{\vec{k}}

\renewcommand{\k}[1]{{\vphantom{k}\smash{k_{#1}}}}
\newcommand{\kp}[1]{{\vphantom{k}\smash{k'_{#1}}}}

\begin{document}

\twocolumn[
  \icmltitle{Tensor Monte Carlo: Particle Methods for the GPU Era}




  \begin{icmlauthorlist}
  \icmlauthor{Laurence Aitchison}{janelia}
  \end{icmlauthorlist}

  \icmlaffiliation{janelia}{Janelia Research Campus, Ashburn, VA 20147, USA}

  \icmlcorrespondingauthor{Laurence Aitchison}{laurence.aitchison@gmail.com}

  \icmlkeywords{VAE, importance sampling, message-passing}

  \vskip 0.3in
]



\printAffiliationsAndNotice{}  

\begin{abstract}
Multi-sample, importance-weighted variational autoencoders (IWAE) give tighter bounds and more accurate uncertainty estimates than variational autoencoders (VAE) trained with a standard single-sample objective.
However, IWAEs scale poorly: as the latent dimensionality grows, they require exponentially many samples to retain the benefits of importance weighting.
While sequential Monte-Carlo (SMC) can address this problem, it is prohibitively slow because the resampling step imposes sequential structure which cannot be parallelised, and moreover, resampling is non-differentiable which is problematic when learning approximate posteriors.
To address these issues, we developed tensor Monte-Carlo (TMC) which gives exponentially many importance samples by separately drawing $K$ samples for each of the $n$ latent variables, then averaging over all $K^n$ possible combinations.
While the sum over exponentially many terms might seem to be intractable, in many cases it can be computed efficiently as a series of tensor inner-products.
We show that TMC is superior to IWAE on a generative model with multiple stochastic layers trained on the MNIST handwritten digit database, and we show that TMC can be combined with standard variance reduction techniques.
\end{abstract}
Variational autoencoders (VAEs) \citep{kingma2013auto,rezende2014stochastic,eslami2018neural} have had dramatic success in exploiting modern deep learning methods to do probabilistic inference in previously intractable high-dimensional spaces.
However, standard VAEs using a single-sample objective give loose variational bounds and poor approximations to the posterior \citep{turner2011two,burda2015importance}.
Modern variational autoencoders instead use a multi-sample objective to improve the tightness of the variational bound and the quality of the approximate posterior \citep{burda2015importance}.
These methods implicitly improve the approximate posterior by drawing multiple samples from a proposal, and resampling to discard samples that do not fit the data \citep{cremer2017reinterpreting}.

While multi-sample importance-weighted methods are often extremely effective, they scale poorly with problem size. 
In particular, recent results \citep{chatterjee2015sample} have shown that the number of importance samples required to closely approximate any target expectation scales as $\exp(\Dkl{\text{P}|| \text{Q}})$, where $\text{Q}$ is the proposal distribution.
Critically, the KL-divergence scales roughly linearly in problem size (and exactly linearly if we consider $n$ independent sub-problems being combined), and thus we expect the required number of importance samples to be exponential in the problem size.
As such, multi-sample methods are typically only used to infer the latent variables in smaller models with only a single (albeit vector-valued) latent variable \citep[e.g.\ ][]{burda2015importance}.

One approach to resolving these issues is sequential Monte-Carlo (SMC) \citep{maddison2017filtering,naesseth2017variational,le2017auto}, which circumvents the need for exponentially many samples using resampling.
(Note that the name TMC is an adaptation of SMC, as the approach is --- at its most general --- a Monte-Carlo method for computing low-variance unbiased estimators).
However, SMC has two issues.
First, the SMC resampling steps force an inherently sequential structure on the computation, which can prohibit effective parallelisation on modern GPU hardware.
While this is acceptable in a model (such as a state-space model) that already has sequential structure, SMC has been applied in many other settings where there is considerably more scope for parallelisation such as mixture models \citep{fearnhead2004particle} or even probabilistic programs \citep{wood2014new}.
Second, modern variational inference uses the reparameterisation trick to obtain low-variance estimates of the gradient of the objective with respect to the proposal parameters \citep{kingma2013auto,rezende2014stochastic}.
However, the reparameterisation trick requires us to differentiate samples from the proposal with respect to parameters of the proposal, and this is not possible in SMC due to the inherently non-differentiable resampling step.
As such, while it may be possible in some circumstances to obtain reasonable results using a biased gradient \citep{maddison2017filtering,naesseth2017variational,le2017auto}, those results are empirical and hence give no guarantees.

To resolve these issues, we introduce tensor Monte Carlo (TMC).
While standard multi-sample objectives draw $K$ samples from a proposal over all latent variables jointly, TMC draws $K$ samples for each of the $n$ latent variables separately, then forms a lower-bound by averaging over all $K^n$ possible combinations of samples for each latent variable.
To perform these averages over an exponential number of terms efficiently, we exploit conditional independence structure in a manner that is very similar to early work on graphical models \citep{pearl1986fusion,lauritzen1988local}.
In particular, we note that for TMC, as well as for classical graphical models, these sums can be written in an extremely simple and general form: as a series of tensor inner products.
This formalism readily allows us to perform exact summation efficiently in difficult cases (e.g. loopy graphs), and can be described graphically as a series of reductions of a factor graph.

Finally, as TMC is, in essence, IWAE with exponentially many importance samples, it can be combined with previously suggested variance reduction techniques, including (but not limited to) sticking the landing (STL) \citep{roeder2017sticking}, doubly reparameterised gradient estimates (DReGs) \citep{tucker2018doubly}, and reweighted wake-sleep (RWS) \citep{bornschein2014reweighted,le2018revisiting}.

\section{Background}

Classical variational inference consists of optimizing a lower bound, $\L_\text{VAE}$, on the log-marginal likelihood, $\log \P{x}$,
\begin{align*}
  \log \P{x} &\geq \L_\text{VAE} = \log \P{x} - \Dkl{\Q{z}|| \P{z| x}},
\end{align*}
where $x$ is the data, $z$ is the latent variable, $\text{P}$ is the generative model, and $\text{Q}$ is known as either the approximate posterior, the recognition model or the proposal distribution.
As the Kullback-Leibler (KL) divergence is always positive, we can see that the objective is indeed a lower bound, and if the approximate posterior, $\Q{z}$, is sufficiently flexible, then as we optimize $\L_\text{VAE}$ with respect to the parameters of the approximate posterior, the approximate posterior will come to equal the true posterior, at which point the KL divergence is zero, so optimizing $\L_\text{VAE}$ reduces to optimizing $\log \P{x}$.
However, in the typical case where $\Q{z}$ is a more restrictive family of distributions, we obtain biased estimates of the generative parameters and approximate posteriors that underestimate uncertainty \citep{minka2005divergence,turner2011two}.

This issue motivated the development of more general lower-bound objectives, and to understand how these bounds were developed, we need to consider an alternative derivation of $\L_\text{VAE}$.
The general approach is to take an unbiased stochastic estimate of the marginal likelihood, denoted $\mathcal{P}$,
\begin{align*}
  \P{x} &= \E[\text{Q}]{\mathcal{P}}
\end{align*}
and convert it into a lower bound on the log-marginal likelihood using Jensen's inequality,
\begin{align*}
  \log \P{x} &\geq \L = \E[\text{Q}]{\log \mathcal{P}}.
\end{align*}
We can obtain most methods of interest, including single-sample VAE's, multi-sample IWAE, and TMC by making different choices for $\mathcal{P}$ and $\text{Q}$.
For the single-sample variational objective we use a proposal, $\Q{z}$, defined over a single setting of the latents,
\begin{align*}
  \mathcal{P}_\text{VAE} &= \frac{\P{x, z}}{\Q{z}},
\end{align*}
which gives rise to the usual variational lower bound, $\L_\text{VAE}$.
However, this single-sample estimate of the marginal likelihood has high variance, and hence a gives a loose lower-bound.
To obtain a tigher variational bound, one approach is to find a lower-variance estimate of the marginal likelihood, and an obvious way to reduce the variance is to average multiple independent samples of the original estimator,
\begin{align*}
  \mathcal{P}_\text{IWAE} &= \frac{1}{K} \sum_{k=1}^K \frac{\P{x, z^k}}{\Q{z^k}},
\end{align*}
which indeed gives rise to a tighter, importance-weighted bound, $\L_\text{IWAE}$ \citep{burda2015importance}.

\section{Results}

First, we give a proof showing that we can obtain unbiased estimates of the model-evidence by dividing the full latent space into several different latent variables, $z = (z_1, z_2,\dotsc,z_n)$, drawing $K$ samples for each individual latent and averaging over all $K^n$ possible combinations of samples.
We then give a method for efficiently computing the required averages over an exponential number of terms using tensor inner products.
We give toy experiments, showing that the TMC bound approches the true model evidence with exponentially fewer samples than IWAE, and in far less time than SMC.
Finally, we do experiments on VAE's with multiple stochastic layers trained on the MNIST handwritten digit database.
We show that TMC can be used to learn recognition models, that it can be combined with variance reduction techniques such as STL \citep{roeder2017sticking} and DReGs \citep{tucker2018doubly}, and is superior to IWAE's given the same number of particles, despite negligable additional computational costs.

\subsection{TMC for factorised proposals}

In TMC we consider models with multiple latent variables, $z = (z_1, z_2,\dotsc,z_n)$, so the generative and recognition models can be written as,
\begin{align*}
  \P{x, z} &= \P{x, z_1, z_2,\dotsc,z_n} \\ 
  \Q{z} &= \Q{z_1}\Q{z_2}\dotsm\Q{z_n}
\end{align*}
where we use a factorised proposal to simplify the proof (see Appendix for a proof for the non-factorised case). 
For the TMC objective, each individual latent variable, $z_i$, is sampled $K_i$ times, 
\begin{align*}
  z_i^{k_i} &\sim \Q{z_i},
\end{align*}
where $i \in \{1,\dotsc,n\}$ indexes the latent variable and $\k{i} \in \{1,\dotsc,K_i\}$ indexes the sample for the $i$th latent.
Importantly, any combination of the $k_i$'s can be used to form an unbiased, single-sample estimate of the marginal likelihood.  Thus, for any $k_1,k_2,\dotsc,k_n$ we have,
\begin{align*}
   \P[\theta]{x} = 
   \E[\Q{z}]{\frac{\P{x, z_1^\k{1}, z_2^\k{2},\dotsc,z_n^\k{n}}}{\Q{z_1^\k{1}}\Q{z_2^\k{2}}\dotsm\Q{z_n^\k{n}}}}.
\end{align*}
The average of a set of unbiased estimators is another unbiased estimator.
As such, averaging over all $K^n$ settings for the $k_i$'s (and hence over $K^n$ unbiased estimators), we obtain a lower-variance unbiased estimator,
\begin{align}
  \label{eq:is}
  \mathcal{P}_\text{TMC} &= \frac{1}{\prod_i K_i} \sum_{\k{1},\k{2},\dotsc,\k{n}} \frac{\P{x, z_1^\k{1}, z_2^\k{2},\dotsc,z_n^\k{n}}}{\Q{z_1^\k{1}}\Q{z_2^\k{2}}\dotsm\Q{z_n^\k{n}}},
\end{align}
and this forms the TMC estimate of the marginal likelihood.

\subsection{Efficient averaging}

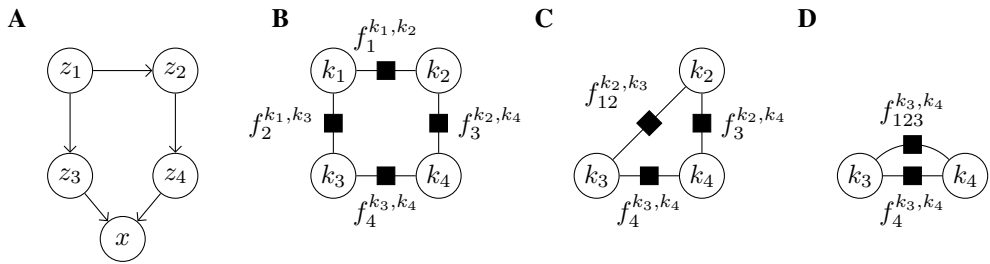
\begin{figure*}
  \centering
  \begin{tikzpicture}
    \tikzstyle{graphnode} = [circle,draw=black,minimum size=17pt,text centered, text width=10pt, inner sep=0pt]
    \tikzstyle{fac}   =[rectangle,draw=black,fill=black,minimum size=5pt]

    \def\d{1.4cm}
    \def\f{0.6cm}
    \def\s{3.5cm}
    \node at ({-0.5*\d}, {0.5*\d}) {\textbf{A}};
    \node[graphnode] (z1) at ({0*\d}, {-0*\d}) {$z_1$};
    \node[graphnode] (z2) at ({1*\d}, {-0*\d}) {$z_2$};
    \node[graphnode] (z3) at ({0*\d}, {-1*\d}) {$z_3$};
    \node[graphnode] (z4) at ({1*\d}, {-1*\d}) {$z_4$};
    \node[graphnode] (x)  at ({0.5*\d}, {-1.6*\d}) {$x$};

    \draw[-{Straight Barb}] (z1) to (z2);
    \draw[-{Straight Barb}] (z2) to (z4);
    \draw[-{Straight Barb}] (z1) to (z3);
    \draw[-{Straight Barb}] (z3) to (x);
    \draw[-{Straight Barb}] (z4) to (x);

    \node at ({-0.5*\d+\s}, {0.5*\d}) {\textbf{B}};
    \node[graphnode] (k1) at ({0*\d+\s}, {-0*\d}) {$k_1$};
    \node[graphnode] (k2) at ({1*\d+\s}, {-0*\d}) {$k_2$};
    \node[graphnode] (k3) at ({0*\d+\s}, {-1*\d}) {$k_3$};
    \node[graphnode] (k4) at ({1*\d+\s}, {-1*\d}) {$k_4$};

    \draw (k1) to node[fac, midway] (f1) {} (k2);
    \draw (k1) to node[fac, midway] (f2) {} (k3);
    \draw (k4) to node[fac, midway] (f3) {} (k2);
    \draw (k4) to node[fac, midway] (f4) {} (k3);

    \node[anchor=south] at (f1.north) {$f_1^{k_1,k_2}$};
    \node[anchor=east]  at (f2.west)  {$f_2^{k_1,k_3}$};
    \node[anchor=west]  at (f3.east)  {$f_3^{k_2,k_4}$};
    \node[anchor=north] at (f4.south) {$f_4^{k_3,k_4}$};

    \node at ({-0.5*\d + 2*\s}, {0.5*\d}) {\textbf{C}};
    \node[graphnode] (k2) at ({1*\d + 2*\s}, {-0*\d}) {$k_2$};
    \node[graphnode] (k3) at ({0*\d + 2*\s}, {-1*\d}) {$k_3$};
    \node[graphnode] (k4) at ({1*\d + 2*\s}, {-1*\d}) {$k_4$};

    \draw (k2) to node[fac, midway, rotate=45] (f12) {} (k3);
    \draw (k4) to node[fac, midway] (f3) {} (k2);
    \draw (k4) to node[fac, midway] (f4) {} (k3);

    \node at ($ (f12) + (  135: \f) $) {$f_{12}^{k_2,k_3}$};
    \node[anchor=west]  at (f3.east)  {$f_3^{k_2,k_4}$};
    \node[anchor=north] at (f4.south) {$f_4^{k_3,k_4}$};

    \node at ({-0.5*\d + 3*\s}, {0.5*\d}) {\textbf{D}};
    \node[graphnode] (k3) at ({0*\d + 3*\s}, {-1*\d}) {$k_3$};
    \node[graphnode] (k4) at ({1*\d + 3*\s}, {-1*\d}) {$k_4$};

    \draw (k3) to[bend left=45]  node[fac, midway] (f123) {} (k4);
    \draw (k3) to node[fac, midway] (f4) {} (k4);

    \node[anchor=south] at (f123.north) {$f_{123}^{k_3,k_4}$};
    \node[anchor=north] at (f4.south) {$f_4^{k_3,k_4}$};

  \end{tikzpicture}
  \caption{
    A graphical depiction of the proceedure for efficiently computing the marginal likelihood for a loopy factor graph.
    \textbf{A.} The original graphical model.
    \textbf{B.} Representing the TMC unbiased estimator as a factor graph.
    \textbf{C.} Summing over $k_1$ simplifies the graph.
    \textbf{D.} Summing over $k_2$ gives a simple graph that can readily be summed out.
    \label{fig:factor:loop}
  }
\end{figure*}

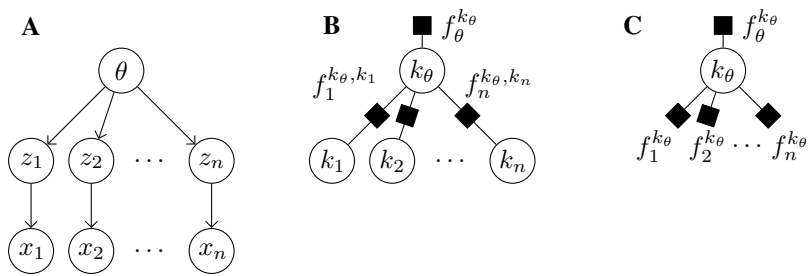
\begin{figure*}
  \centering
  \begin{tikzpicture}
    \tikzstyle{graphnode} = [circle,draw=black,minimum size=17pt,text centered, text width=10pt, inner sep=0pt]
    \tikzstyle{fac}   =[rectangle,draw=black,fill=black,minimum size=5pt]

    \def\d{1.2cm}
    \def\f{0.6cm}
    \def\s{4cm}
    \node at (0, {0.5*\d}) {\textbf{A}};
    \node[graphnode] (theta) at (\d, 0) {$\theta$};
    \node[graphnode] (z1) at ({  0*\d}, {-1*\d}) {$z_1$};
    \node[graphnode] (z2) at ({2/3*\d}, {-1*\d}) {$z_2$};
    \node            (zd) at ({4/3*\d}, {-1*\d}) {$\dots$};
    \node[graphnode] (zn) at ({  2*\d}, {-1*\d}) {$z_n$};

    \node[graphnode] (x1) at ({  0*\d}, {-2*\d}) {$x_1$};
    \node[graphnode] (x2) at ({2/3*\d}, {-2*\d}) {$x_2$};
    \node            (xd) at ({4/3*\d}, {-2*\d}) {$\dots$};
    \node[graphnode] (xn) at ({  2*\d}, {-2*\d}) {$x_n$};

    \draw[-{Straight Barb}] (theta) to (z1);
    \draw[-{Straight Barb}] (theta) to (z2);
    \draw[-{Straight Barb}] (theta) to (zn);
    \draw[-{Straight Barb}] (z1) to (x1);
    \draw[-{Straight Barb}] (z2) to (x2);
    \draw[-{Straight Barb}] (zn) to (xn);

    \node at (\s, {0.5*\d}) {\textbf{B}};
    \node[graphnode] (theta) at ({\d + \s}, 0) {$k_{\theta}$};
    \node[graphnode] (z1) at ({  0*\d+\s}, {-1*\d}) {$k_1$};
    \node[graphnode] (z2) at ({2/3*\d+\s}, {-1*\d}) {$k_2$};
    \node            (zd) at ({4/3*\d+\s}, {-1*\d}) {$\dots$};
    \node[graphnode] (zn) at ({  2*\d+\s}, {-1*\d}) {$k_n$};

    \node[fac] (ftheta) at ({\d+\s}, {0.5*\d}) {};
    \draw (ftheta) to (theta);
    \draw (theta) to node[fac, midway, rotate=45] (f1) {} (z1);
    \draw (theta) to node[fac, midway, rotate=-18.4] (f2) {} (z2);
    \draw (theta) to node[fac, midway, rotate=45] (fn) {} (zn);

    \node[anchor=west] at (ftheta.east) {$f_{\theta}^{k_\theta}$};
    \node at ($ (f1) + (  135: \f) $) {$f_{1}^{k_\theta,k_1}$};
    \node at ($ (fn) + (   45: \f) $) {$f_{n}^{k_\theta,k_n}$};

    \node at ({2*\s}, {0.5*\d}) {\textbf{C}};
    \node[graphnode] (theta) at ({\d + 2*\s}, 0) {$k_{\theta}$};

    \node[fac] (ftheta) at ({\d+2*\s}, {0.5*\d}) {};

    \node[fac,rotate= 45  ] (f1) at ({1/2*\d+2*\s}, {-1/2*\d}) {};
    \node[fac,rotate=-18.4] (f2) at ({5/6*\d+2*\s}, {-1/2*\d}) {};
    \node[fac,rotate= 45  ] (fn) at ({3/2*\d+2*\s}, {-1/2*\d}) {};

    \draw (ftheta) to (theta);
    \draw (f1) to (theta);
    \draw (f2) to (theta);
    \draw (fn) to (theta);

    \node[anchor=west] at (ftheta.east) {$f_{\theta}^{k_\theta}$};
    \node at ($ (f1) + (-0.3cm, -0.4cm) $) {$f_{1}^{k_\theta}$};
    \node at ($ (f2) + ( 0.0cm, -0.4cm) $) {$f_{2}^{k_\theta}$};
    \node at ($ (fn) + ( 0.3cm, -0.4cm) $) {$f_{n}^{k_\theta}$};

    \node at ({7.75/6*\d+2*\s}, {-1/2*\d-0.4cm}) {$\dots$};

  \end{tikzpicture}
  \caption{
    A graphical depiction of the proceedure for efficiently computing the marginal likelihood for a latent variable model with unknown parameters, $\theta$, and latents, $z_i$ corresponding to each data point, $x_i$.
    \textbf{A.} The original graphical model.
    \textbf{B.} Representing the TMC unbiased estimator as a factor graph.
    \textbf{C.} Summing over $k_1,k_2,\dotsc k_n$ simplifies the graph, allowing the TMC estimator to be readily computed by summing over $k_\theta$.
    \label{fig:factor:lvm}
  }
\end{figure*}

The TMC unbiased estimator in Eq.~\eqref{eq:is} involves a sum over exponentially many terms, which may be intractable.
To evaluate the TMC marginal likelihood estimate efficiently, we therefore need to exploit structure in the graphical model.
For instance, for a directed graphical model, the joint-probability can be written as a product of the conditional probabilities,
\begin{align*}
  \P{x, z_1, z_2,\dotsc,z_n} &= \P{x| z_{\text{pa}(x)}}\prod_{i=1}^n \P{z_i| z_{\text{pa}(z_i)}}
\end{align*}
where $\text{pa}(z_i) \subset \{1,\dotsc,n\}$ is the indicies of the parents of $z_i$.
In this (and many other cases), we can write the importance ratio as a product of factors,
\begin{align}
  \label{eq:fac}
  \prod_j f_j^{\kappa_j} &= \frac{\P{x, z_1^\k{1}, z_2^\k{2},\dotsc,z_n^\k{n}}}{\Q{z_1^\k{1}}\Q{z_2^\k{2}}\dotsm\Q{z_n^\k{n}}}
\end{align}
where, $\kappa_j$'s are tuples containing the indicies ($k_i$'s) of each factor (see Appendix for further details).
To average over all possible combinations of samples for each latent variable, we use,
\begin{align*}
  \mathcal{P}_\text{TMC} &= \frac{1}{K_1 K_2 \dotsm K_n} \sum_{k_1, k_2,\dotsc,k_n} \prod_{j} f_j^{\kappa_j}.
\end{align*}
If there are sufficiently many conditional independencies in the graphical model, we can compute the TMC marginal likelihood estimate efficiently by swapping the order of the product and summation.

For instance, consider the generative model in Fig.~\ref{fig:factor:loop}A.
The corresponding TMC estimator is
\begin{align*}
  \mathcal{P}_\text{TMC} &= \frac{1}{K_1 K_2 K_3 K_4} \sum_{k_1,k_2,k_3,k_4} f_1^{k_1 k_2} f_2^{k_1 k_3} f_3^{k_2 k_4} f_4^{k_3 k_4},
\end{align*}
which can be understood by reference to a loopy factor graph defined over $k_1,k_2,k_3,k_4$ (Fig.~\ref{fig:factor:loop}B). 
Summing over $k_1$, we obtain Fig.~\ref{fig:factor:loop}C,
\begin{align*}
  \mathcal{P}_\text{TMC} &= \frac{1}{K_2 K_3 K_4} \sum_{k_2,k_3,k_4} f_{12}^{k_2 k_3} f_3^{k_2 k_4} f_4^{k_3 k_4} \\
  f_{12}^{k_2 k_3} &= \frac{1}{K_1} \sum_{k_1} f_1^{k_1 k_2} f_2^{k_1 k_3},
  \intertext{and summing over $k_2$ we obtain Fig.~\ref{fig:factor:loop}D,}
  \mathcal{P}_\text{TMC} &= \frac{1}{K_3 K_4} \sum_{k_3,k_4} f_{123}^{k_3 k_4} f_4^{k_3 k_4} \\ 
  f_{123}^{k_3 k_4} &= \frac{1}{K_2} \sum_{k_2} f_{12}^{k_2 k_3} f_3^{k_2 k_4},
\end{align*}
which be computed directly.
Now we can find the optimal settings for the generative and proposal parameters by performing gradient ascent on $\log \mathcal{P}_\text{TMC}$ using standard automatic differentation tools.

As a second more practical example, consider the generative model in Fig.~\ref{fig:factor:lvm}A, with unknown parameters, $\theta$, and unknown latents, $z_i$, corresponding to each data point, $x_i$.
The corresponding TMC estimator is,
\begin{align*}
  \mathcal{P}_\text{TMC} &= \frac{1}{K_\theta K^N} \sum_{k_\theta,k_1,k_2,\dotsc,k_N} f_\theta^{k_\theta} \prod_{i=1}^N f_i^{k_\theta, k_i},
\end{align*}
where $k_i$, which runs from $1$ to $K$, indexes samples of $z_i$ and $k_\theta$, which runs from $1$ to $K_\theta$, indexes samples of $\theta$.
We can represent this estimator as a factor graph (Fig.~\ref{fig:factor:lvm}B).
To efficiently compute the TMC estimate, we sum over $k_1,k_2,\dotsc,k_n$,
\begin{align*}
  \mathcal{P}_\text{TMC} &= \frac{1}{K_\theta} \sum_{k_\theta=1}^{K_\theta} f_\theta^{k_\theta} \prod_{i=1}^N f_i^{k_\theta} \\ 
  f_i^{k_\theta} &= \frac{1}{K} \sum_{k_i=1}^K f_i^{k_\theta, k_i},
\end{align*}
which is represented in Fig.~\ref{fig:factor:lvm}C and can be computed directly.

\subsection{Non-factorised proposals}
For non-factorised proposals, we obtain a result similar to that for factorised proposals (Eq.~\ref{eq:is}),
\begin{align}
  \mathcal{P}_\text{TMC} &= \frac{1}{\prod_i K_i} \sum_{\k{1},\k{2},\dotsc,\k{n}} \frac{\P{x, z_1^\k{1}, z_2^\k{2},\dotsc,z_n^\k{n}}}{\prod_i \Q{z_i^\k{i}| \z_\text{qa}(z_i)}}
\end{align}
where $\text{qa}(z_i)$ represents the parents of $z_i$ under the proposal, and $\z_\text{qa}(i)$ represents all samples of those parents.
Importantly, note that the proposals are indexed only by $k_i$, and not by $k_{\text{qa}(z_i)}$, so we can always use the same factorisation structure (Eq.~\ref{eq:fac}) for a factorised and non-factorised proposal.
Consult the appendix for further details, including a proof.

\subsection{Computational costs for TMC and IWAE}

In principle, TMC could be considerably more expensive than IWAE, as IWAE's cost is linear in $K$, whereas for TMC, the cost scales with $K^{\operatorname{max}(\lvert\kappa\rvert)}$, where $\operatorname{max}(\lvert\kappa\rvert)$ denotes largest number of indicies required for a factor, $f^{\kappa_j}_j$.
Of course, in exchange, we obtain an exponential number of importance samples, $K^n$, so this tradeoff will usually be worthwhile.
Remarkably however, in many modern deep models, the computational cost of TMC is \textit{linear} in $K$, and almost equivalent to that of IWAE.
In particular, consider a chained model, where $z = (z_1, z_2, \dotsc, z_n)$, and,
\begin{align*}
  \P{x, z} &= \P{x| z_1} \P{z_1| z_2} \dotsm \P{z_{n-1}| z_n} \P{z_n}
\end{align*}
In most deep models, the latents, $z_i$, are vectors, and the generative (and recognition) models have the form,
\begin{align*}
  \P{z_{i}| z_{i+1}} &= \N{z_{i}| \mu_i(z_{i+1}), \text{diag}(\sigma_i^2(z_{i+1}))}
\end{align*}
i.e.\ the elements of $z_i$ are independent, with means and variances given by neural-networks applied to the activations of the previous layer, $\mu_i(z_{i+1})$ and $\sigma_i^2(z_{i+1})$.
As such, the nominally quadratic cost of evaluating $\P{z_i^\k{i}| z_{i+1}^\k{i+1}}$ for all $\k{i}$ and $\k{i+1}$ is dominated by the \textit{linear} cost of computing $\mu_i(z_{i+1}^\k{i+1})$ and $\sigma_i^2(z_{i+1}^\k{i+1})$ for all $\k{i+1}$ by applying neural networks to the activations at the previous layer.
Finally, note that recognition models are generally defined in a similar way \citep[e.g.\ ][]{kingma2013auto,rezende2014stochastic,burda2015importance,sonderby2016ladder}, and as such, the corresponding recognition models will also usually have linear cost.

\section{Toy Experiments}
Here we perform two toy experiments.
First, we compare TMC, SMC and IWAE, finding that TMC gives bounds on the log-probability that are considerably better than those for IWAE, and as good (if not better than) SMC, while being considerably faster.
Second, we consider an example where non-factorised posteriors might become important.
In these toy experiments, we use models in which all variables are jointly Gaussian, which allows us to compute the exact marginal likelihood, and to assess the tightness of the bounds.

\subsection{Comparing TMC, SMC and IWAE}
\begin{figure*}
  \centering
  \includegraphics{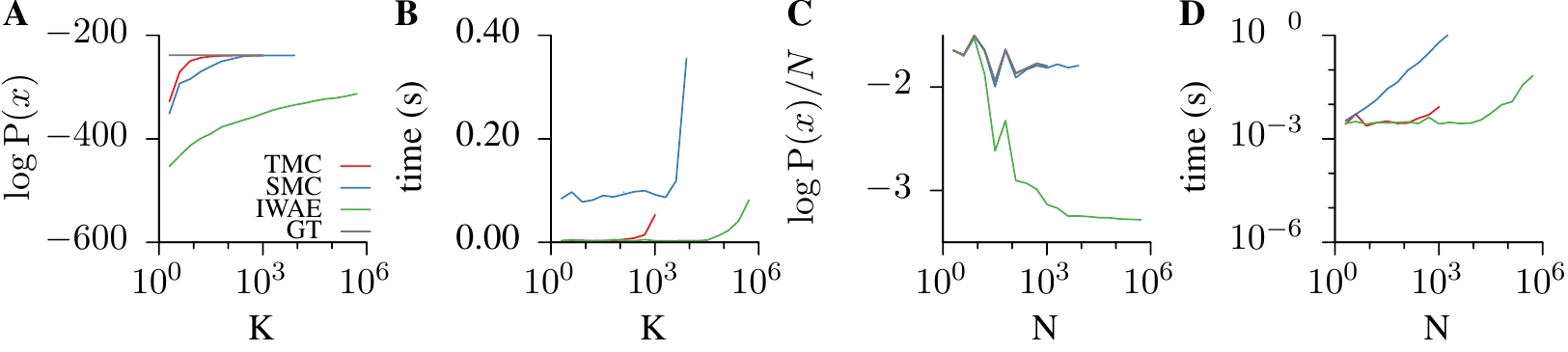}
  \caption{
    Performance of TMC, SMC, IWAE and ground truth (GT)  on a simple Gaussian latent variable example, run in PyTorch using a GPU.
    \textbf{A.} The marginal likelihood estimate (y-axis) for different numbers of particles, $K$ (x-axis), with the number of data points fixed to $N=128$.
    \textbf{B.} The time required for computing marginal likelihood estimates in \textbf{A} on a single Titan X GPU.
    \textbf{C.} The marginal likelihood estimate per data point (y-axis), for models with different numbers of data points, $N$, and a fixed number of particles, $K=128$.  Note that the TMC, SMC and GT lines lie on top of each other.
    \textbf{D.} The time required for computing marginal likelihood estimates in \textbf{C}.
    \label{fig:param}
  }
\end{figure*}

First, we considered a simple example, with Gaussian parameters, latents and data. 
There was a single parameter, $\theta$, drawn from a standard normal, which set the mean of $N$ latent variables, $z_i$. 
The $N$ data points, $x_i$, have unit variance, and mean set by the latent variable,
\begin{align*}
  \P{\theta} &= \N{\theta; 0, 1}, \\
  \P{z_i| \theta} &= \N{z_i; \theta, 1}, \\
  \P{x_i| z_i} &= \N{x_i; z_i, 1}.
\end{align*}
For the proposal distributions for all methods, we used the generative marginals,
\begin{align*}
  \Q{\theta} &= \N{\theta; 0, 1}, \\
  \Q{z_i} &= \N{z_i; 0, 2}.
\end{align*}
While this model is simplistic, it is a useful initial test case, because the ground true marginal likelihood can be computed (GT).

We computed marginal likelihood estimates for TMC, SMC and IWAE.
First, we plotted the bound on the marginal likelihood against the number of particles for a fixed number of data points, $N=128$ (Fig.~\ref{fig:param}A).
As expected, IWAE was dramatically worse than other methods: it was still far from the true marginal likelihood estimate, even with one million importance samples.
We also found that for a fixed number of particles/samples, TMC was somewhat superior to SMC. 
We suspect that this is because TMC sums over all possible combinations of particles, while the SMC resampling step explicitly eliminates some of these combinations.
Further, SMC was considerably slower than TMC as resampling required us to perform an explicit loop over data points, whereas TMC can be computed entirely using tensor sums/products, which can be optimized efficiently on the GPU (Fig.~\ref{fig:param}B).

Next, we plotted the log-marginal likelihood per data point as we vary the number of data points, with a fixed number of importance samples, $K=128$ (Fig.~\ref{fig:param}C).
Again, IWAE is dramatically worse than the other methods, whereas both TMC and SMC closely tracked the ground-truth result.
However, note that the time taken for SMC (Fig.~\ref{fig:param}D) is larger than that for the other methods, and scales linearly in the number of data points.
In contrast, the time required for TMC remains constant up to around $1000$ data points, as GPU parallelisation is exploited increasingly efficiently in larger problems. 

\subsection{Comparing factorised and non-factorised proposals}

Non-factorised proposals have a range of potential benefits, and here we consider how they might be more effective than factorised proposals in modelling distributions with very high prior correlations.
In particular, we consider a chain of latent variables,
\begin{align*}
  \P{z_i| z_{i-1}} &= \N{z_{i-1}, 1/N}, \\ 
  \P{x| z_N} &= \N{z_N, 1},
\end{align*}
where $z_0 = 0$.  As $N$ becomes large, the marginal distribution over $z_N$ and $x$ remains constant, but the correlations between adjacent latents (i.e.\ $z_{i-1}$ and $z_i$) become stronger.
For the factorised proposal, we use the marginal variance (i.e.\ $\Q{z_i}=\N{0, i/N}$), whereas we used the prior for the non-factorised proposal.
Taking $N=100$, we find that the non-factorised method considerably outperforms the factorised method for small numbers of samples, $K$, because the non-factorised method is able to model tight prior-induced correlations.

\begin{figure}
  \centering
  \includegraphics{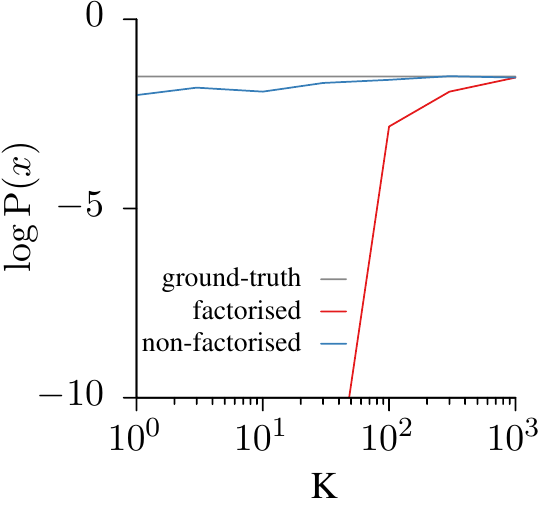}
  \caption{
    The bound on the log-marginal likelihood for factorised and non-factorised models as compared to the ground-truth, for different number of samples, $K$.
    \label{fig:non-fac}
  }
\end{figure}

\section{Experiments}
\begin{figure*}[t]
  \centering
  \includegraphics{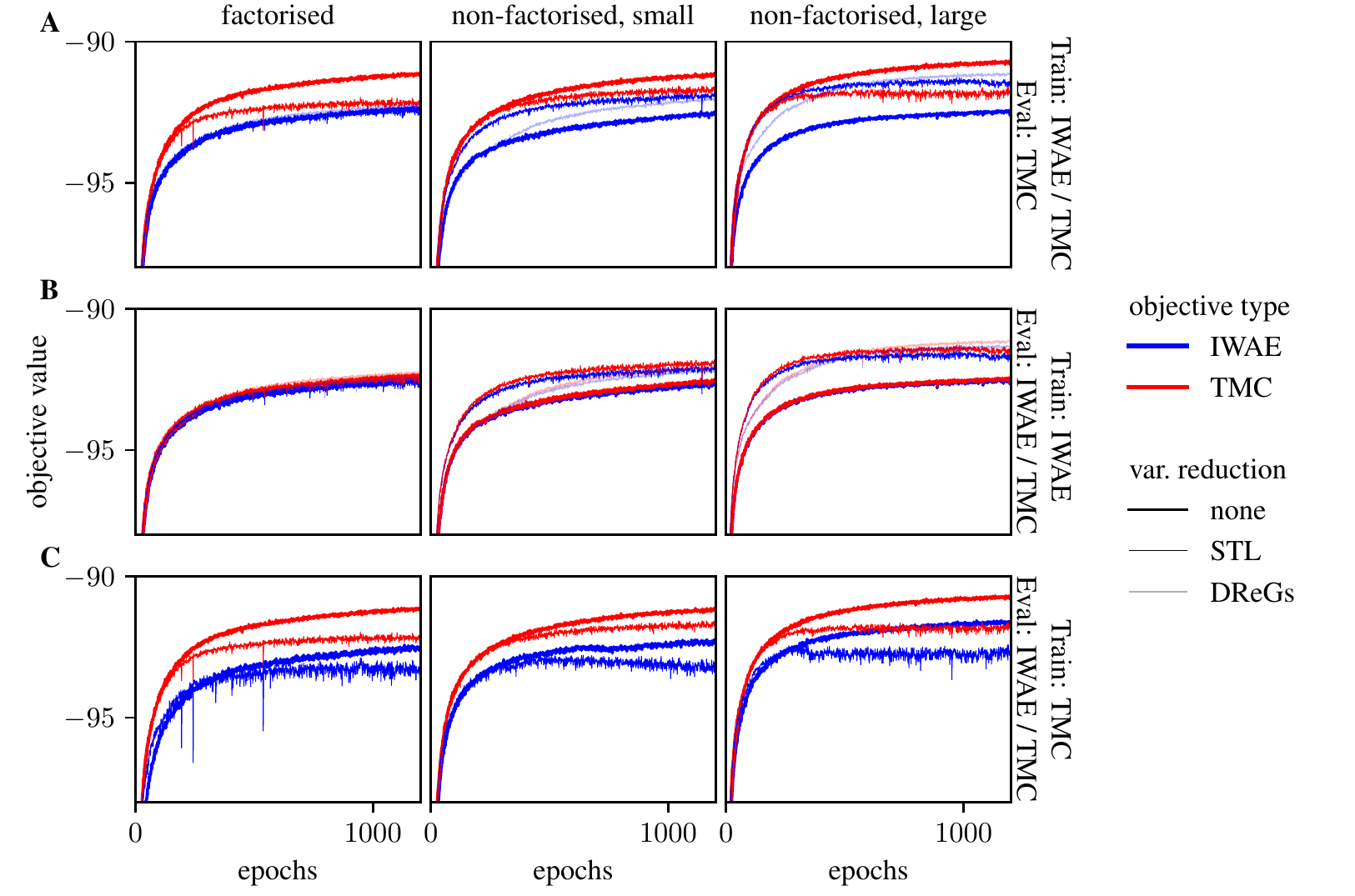}
  \caption{
    The quality of the variational lower bound for a model of MNIST handwritten digits, with different recognition models and training schemes. 
    We used three different recognition models (columns): factorised (left) where the distribution over the latents at each layer was independent; non-factorised, small (middle) where each stochastic layer depended on the previous stochastic layer through a two-layer deterministic neural network, with a small number of units (the same as in the generative model); and non-factorised, large (right) where the deterministic networks linking stochastic layers in the recognition model had 4 times as many units as in the smaller network.
    \textbf{A.} We trained two sets of models using IWAE (blue) and TMC (red), and plotted the value of the TMC objective for both lines.
    \textbf{B.} Here, we consider only models trained using the IWAE objective, and evalute them under the IWAE objective (blue) and the TMC objective (red).
    \textbf{C.} Here, we consider only models trained using the TMC objective, and evalute them under the IWAE objective (blue) and the TMC objective (red).
    \label{fig:vae}
  }
\end{figure*}

We considered a model for MNIST handwritten digits with five layers of stochastic units inspired by \citet{sonderby2016ladder}. 
This model had 4 stochastic units in the top layer (furthest from the data), then 8, 16, 32, and 64 units in the last layer (closest to the data).
In the generative model, we had two determinstic layers between each pair of stochastic layers, and these deterministic layers had twice the number of units in the corresponding stochastic layer (i.e. $8$, $16$, $32$, $64$ and $128$). 
In all experiments, we used the Adam optimizer \citep{kingma2014adam} using the PyTorch default hyperparameters, and weight normalization \citep{salimans2016weight} to improve numerical stability.
We used leaky-relu nonlinearities everywhere except for the standard-deviations \citep{sonderby2016ladder}, for which we used $0.01 + \text{softplus}(x)$, to improve numerical stability by ensuring that the standard deviations could not become too small.
Note, however, that our goal was to give a fair comparison between IWAE and TMC under various variance reduction schemes, not to reach state-of-the-art performance.
As such, there are many steps that could be taken to bring results towards state-of-the-art, including the use of a ladder-VAE architecture, wider deterministic layers, batch-normalization, convolutional structure and using more importance samples to evaluate the model \citep{sonderby2016ladder}.

We compared IWAE and TMC under three different recognition models, as well as three different variance reduction schemes (including plain reparameterisation gradients).
For the non-factorised recognition models (Fig.~\ref{fig:vae} middle and right), we used,
\begin{align*}
  \Q{z| x} &= \Q{z_5| z_4} \Q{z_4| z_3} \Q{z_3| z_2} \Q{z_2| z_1} \Q{z_1| x}.
\end{align*}
For all of these distributions, we used,
\begin{subequations}
\begin{align}
  \label{eq:Q:arch:1}
  \Q{z_{i+1}| z_i} &= \N{\nu_{i+1}, \rho^2_{i+1}},\\
  \label{eq:Q:arch:2}
  \nu_{i+1} &= \text{Linear}(h_i),\\
  \label{eq:Q:arch:3}
  \rho_{i+1} &= \text{SoftPlus}\b{\text{Linear}(h_i)},\\
  \label{eq:Q:arch:4}
  h_i &= \text{MLP}(z_i).
\end{align}
\end{subequations}
where the MLP had two dense layers and for the small model (middle), the lowest-level MLP had 128 units, then higher-level MLPs had 64, 32, 16 and 8 units respectively.
For the large model, the lowest-level MLP had 512 units, then 256, 128, 64 and 32 units in higher level MLP's.
For the factorised recognition model, 
\begin{align}
  \Q{z| x} &= \Q{z_1| x} \Q{z_2| x} \Q{z_3| x} \Q{z_4| x} \Q{z_5| x},
\end{align}
we used an architecture mirroring that of the non-factorised recognition model as closely as possible.
In particular, to construct these distributions, we used Eqs.~(\ref{eq:Q:arch:1}--\ref{eq:Q:arch:3}), but with a different version of $h_i$ that depended directly on $h_{i-1}$, rather than $z_i$,
\begin{align}
  h_i &= \text{MLP}\b{\text{Linear}\b{h_{i+1}}}
\end{align}
where we require the linear transformation to reduce the hidden dimension down to the required input dimension for the MLP.

For IWAE and TMC, we considered plain reparametrised gradient descent (none), as well as two variance reduction techniques: STL \citep{roeder2017sticking}, and DReGs \cite{tucker2018doubly}.

We began by training the above models and variance reduction techniques using an IWAE (blue) and a TMC (red) objective (Fig.~\ref{fig:vae}A).
Note that we evaluated both models using the TMC objective to be as generous as possible to IWAE (see Fig.~\ref{fig:vae}B and discussion below).
We found that the best performing method for all models was plain TMC (i.e. without STL or DReGs; Fig.~\ref{fig:vae}A).
It is unsuprising that TMC is superior to IWAE, because TMC in effect considered $20^5$ importance samples, whereas IWAE considered only $20$ importance samples.
However, it is unclear which variance reduction technique should prove most effective in combination with TMC.
We can speculate that STL and DReGs perform poorly in combination with TMC because these methods are designed to improve performance as the approximate posterior becomes close to the true posterior.
However, TMC considers all combinations of samples, and while some of those combinations might be drawn from the true posterior (e.g.\ we might have all latents for a particular $k$, $z_1^k, z_2^k,\dotsc,z_n^k$, being drawn jointly from the true posterior), it cannot be the case that all combinations are (or can be considered as) drawn jointly from the true posterior.
Furthermore, note that STL is known to be biased, and we found that DReGs was numerically unstable in combination with TMC, though it is not clear whether this was an inherent property or merely an implementational issue.
Furthermore, note that TMC offers larger benefits over IWAE for the factorised and small non-factorised model, where --- presumably --- the mismatch between the approximate and true posterior is larger.
Next, we considered training the model under just IWAE, and evaluating under IWAE and TMC (Fig.~\ref{fig:vae}B).
We found that evaluating under TMC consistently gave a slightly better bound than evaluating under IWAE, despite the model being trained under the IWAE objective.
As such, in Fig.~\ref{fig:vae}A, we used TMC to evaluate the model trained under an IWAE objective, so as to be as generous as possible to IWAE.
Finally, we considered training the model under just TMC and evaluating under IWAE and TMC (Fig.~\ref{fig:vae}C).
We found that there was a dramatic difference between these two evaluation methods, indicating that models trained under the TMC objective do exploit the additional flexibility provided by TMC that is absent under IWAE.

\begin{figure}[t]
  \centering
  \includegraphics{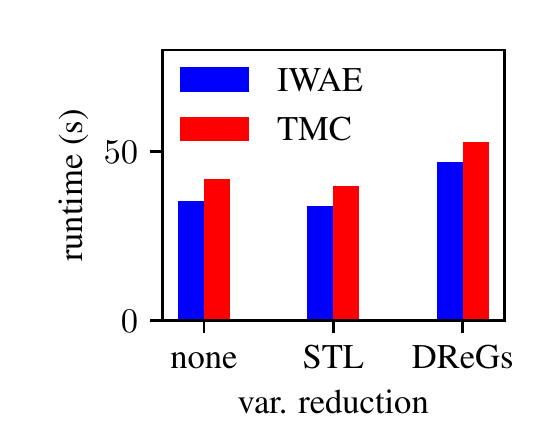}
  \caption{
    The average time (across the three models) required for one training epoch of the six methods considered above: IWAE/TMC in combination with no additional variance reduction scheme (none), STL, DReGs.
    \label{fig:time}
  }
\end{figure}

We found that the training time for TMC was similar to that for IWAE, despite TMC considering --- in effect --- $20^5 = 3,200,000$ importance samples, whereas IWAE considered only $20$ (Fig.~\ref{fig:time}).
Further, we found that using STL gave similar runtime, whereas DReGs was more expensive (though it cannot be ruled out that this is due to our implementation).
That said, broadly, there is reason to believe that vanilla reparameterised gradients may be more efficient than STL and DReGs, because using vanilla reparameterised gradients allows us to compute the recognition sample and log-probability in one pass.
In contrast, for STL and DReGs, we need to separate the computation of the recognition sample and log-probability, so that it is possible to stop gradients in the appropriate places.

\section{Future work}
In future work we hope that TMC will find applications in a broad range of areas, and that ideas from classical probabilistic methods can be used to improve TMC.

There are two particularly striking possible applications for TMC.
First, TMC is potentially very well-suited to probabilistic programming \citep{wingate2011lightweight,wood2014new,mansinghka2014venture}, because it combines the advantages of SMC and variational inference.
In particular, just like SMC it can leverage poor proposals (even sampling from the prior) to obtain reasonable posteriors, but retains the advantages of VAE's, in particular, the ability to use the reparameterisation trick to efficiently learn effective proposal distributions, which is important when there are high-dimensional latents without exploitable conditional independencies.
Second, obtaining interpretable, high-level representations of input data remains an important challenge for deep learning.
One approach to learning such representations is to impose structure, in the form of conditional independencies \citep[e.g.\ ][]{johnson2016composing}, and TMC should aid in the development of such methods by providing an efficient inference method that exploits these conditional independencies.

To improve TMC, there are a variety of directions to consider.
First, it may be possible to improve TMC proposals by exploiting the rich array of methods from SMC, from implementing a particle filter as a proposal to using systematic and stratified resampling \citep{kitagawa1996monte,douc2005comparison}.
Second, TMC as described here is only suitable for sampling distributions with a finite number of latent variables, and not for distributions such as a Dirichlet process with an infinite number of latent variables.
Implementing methods to handle such distributions is important, especially in the probabilistic programming context, and may be achieved by taking inspiration from sampling-based methods for handling such distributions.
Fourth, methods have been developed to optimize discrete proposals specifically in the context of IWAE \citep{mnih2016variational}, and similar methods may be extremely efficient in the context of TMC.

\section{Discussion}

We showed that it is possible to extend multi-sample bounds on the marginal likelihood by drawing samples for each latent variable separately, and averaging across all possible combinations of samples from each variable.
As such, we were able to achieve lower-variance estimates of the marginal likelihood, and hence better bounds on the log-marginal likelihood than IWAE and SMC.
Furthermore, computation of these bounds parallelises effectively on modern GPU hardware, and is comparable to the computation required for IWAE.

Our approach can be understood as introducing ideas from classical message passing \citep{pearl1982reverend,pearl1986fusion,bishop2006pattern} into the domains of importance sampling and variational autoencoders.
Note that while message passing has been used in the context of variational autoencoders to sum over discrete latents \citep[e.g.\ ][]{johnson2016composing}, here we have done something fundamentally different.
In particular we introduce message-passing like approaches into the fabric of importance sampling, by first drawing $K$ samples for each latent, including continuous latents which have no conjugacy properties, and then using message-passing like algorithms to sum over all possible combinations of samples.

\bibliography{refs}
\bibliographystyle{icml2019}

\pagebreak
\newpage

\section{Appendix}
\subsection{Typical choice of factors for a directed graphical model}
Here, we give details of a typical factorisation, as it might follow a directed graphical model.
In particular, we use, one factor corresponding to each variable (latent or observed).
For the latent variables, we combine the prior and proposal into a single factor,
\begin{align*}
  f_j^{\kappa_j} &= \begin{cases} 
    \P{x| z^{k_{\text{pa}(x)}}_{\text{pa}(x)}} & \text{for } j = 0\\[10pt]
    \dfrac{\P{z_j^{k_j}| z^{k_{\text{pa}(z_j)}}_{\text{pa}(z_j)}}}{\Q{z^{k_j}_j}} & \text{for } j \in \{1,\dotsc,n\}
  \end{cases}
\end{align*}
where $k_{\text{pa}(z_j)}$ gives the indicies for all the parents of $z_j$, and $\kappa_j$ gives all the required indicies for the factor,
\begin{align}
  \kappa_j = \begin{cases}
    k_{\text{pa}(x)} &\text{for } 0=j\\[5pt]
    \b{k_j, k_{\text{pa}(z_j)}} &\text{for } j \in \{1,\dotsc,n\} .
  \end{cases}
\end{align}
It is also straightforward to extend this factorisation by considering multiple observed variables, $x = (x_1, x_2, \dotsc, x_m)$.


\subsection{TMC for non-factorised proposals}
\label{sec:tmc}

Now that we have established the possibility of efficiently computing the TMC marginal likelihood estimate, we come back to show that it is possible to use non-factorised proposals in TMC.

Unfortunately, the proof is considerably more involved than the previous proof for factorised TMC, requiring us to consider the joint distribution over all samples for all latents, $\z=(\z_1, \z_2,\dotsc,\z_n)$, where all samples for the $i$th latent are given by $\z_i=(z_i^1,z_i^2,\dotsc,z_i^{K_i})$.
Our approach is to define sets of generative distributions, $\P[\kv]{\z}$ and $\P[\kv]{x| \z}$, indexed by $\kv=(\k{1}, \k{2},\dotsc,\k{n})$ such that for all choices of $\kv$, the usual importance ratio gives an unbiased estimate of the model evidence,
\begin{align}
  \label{eq:non-fac-ref}
  \P{x} &= \E[\Q{\z}]{\P[\kv]{x| \z} \frac{\P[\kv]{\z}}{\Q{\z}}}
\end{align}
To obtain this equality, we split the full latent space, $\z$ into the ``indexed'' latents, $z^\kv = (z_1^\k{1}, z_2^\k{2},\dotsc,z_n^\k{n})$, and the other, ``non-indexed'' latents, $z^{-\kv}$.
We chose the likelihood, $\P[\kv]{x| \z}$, such that the data depends on only the indexed latents, $z^\kv$, in exactly the same way as in the original model,
\begin{align}
  \label{eq:non-fac:likelihood}
  \P[\kv]{x| \z} &= \P{x| z=z^\kv}.
\end{align}
For the prior, we begin by factorising it into terms for the indexed and non-indexed latents,
\begin{align}
  \label{eq:non-fac:prior}
  \P[\kv]{\z} &=  \P{z^{-\kv}| z^\kv} \P{z^\kv}.
  \intertext{and we chose the distribution over the indexed latents to be that under the original model,}
  \P{z^\kv} &= \P{z=z^\kv}.
\end{align}
These two choices are all that is required to give an unbiased estimator of the original model evidence.  In particular,
\begin{multline*}
  \E[\Q{\z}]{\P[\kv]{x| \z} \frac{\P[\kv]{\z}}{\Q{\z}}} =\\ \int dz^\kv \; dz^{-\kv} \; \P{x| z^\kv} \P{z^\kv} \P{z^{-\kv}| z^\kv}
\end{multline*}
integrating over $z^{-\kv}$, then using our choices for the likelihood and prior,
\begin{align*}
  \E[\Q{\z}]{\P[\kv]{x| \z} \frac{\P[\kv]{\z}}{\Q{\z}}} &= \int dz^\kv \; \P{x| z^\kv} \P{z^\kv} \\
  &= \int dz \; \P{x| z} \P{z} = \P{x},
\end{align*}
as required.
Importantly, note that this derivation made no assumptions about the proposal, $\Q{\z}$, and the generative model for the non-indexed latents, $\P{z^{-\kv}| z^\kv}$, giving us complete freedom --- at least in principle --- about how we choose those quantities.

However, importance sampling over the enlarged latent space ($\z$) may give rise to higher variance estimators than working in the orignal space, ($z$ or $z^\kv$).
As such, it pays to be careful about the choice of generative model for the non-indexed latents, $\P{z^{-\kv}| z^\kv}$, and the proposal, $\Q{\z}$.
In particular, our strategy is to choose $\P{z^{-\kv}| z^\kv}$ such that it cancels many of the terms in $\Q{\z}$.
%
We begin by assuming that each sample for a single latent is independent, conditioned on all samples of previous latents,
\begin{align}
  \label{eq:non-fac:proposal}
  \Q{\z| x} &= \prod_i \prod_{k_i} \Q{z_i^{k_i}| x, \z_{\text{qa}(i)}}
\end{align}
where $\text{qa}(z_i) \subseteq \{1,\dotsc,n\}$ gives the indices of the parents of $z_i$ under the proposal.
To give as much cancellation as possible, we assume that the generative model for the non-indexed latents is equal to the proposal,
\begin{align}
  \label{eq:non-fac:non-ind}
  \P{z^{-\kv}| z^\kv} &= \prod_i \prod_{\kp{i} \neq \k{i}} \Q{z_i^{\kp{i}}| x, \z_{\text{qa}(i)}}
\end{align}
After cancelling $\P{z^{-\kv}| z^\kv}$ with terms in the proposal the importance ratio becomes,
\begin{align}
  \P[\kv]{x| \z} \frac{\P[\kv]{\z}}{\Q{\z| x}} &= \frac{\P{x| z^\kv} \P{z^\kv}}{\prod_i \Q{z_i^{k_i}| x, \z_{\text{qa}(i)}}}\\
  &= \frac{\P{x, z_1^\k{1}, z_2^\k{2}, \dotsc, z_n^\k{n}}}{\prod_i \Q{z_i^{k_i}| x, \z_{\text{qa}(i)}}}
\end{align}
Notably, this is analogous to the factorised case, except that the proposal is allowed to depend on samples of the other latents.
Further, note that the techniques for efficient averaging continue to work in exactly the same way: the proposal factors depend only on one index, $\k{i}$, and so we can always use the same factorisation of the importance ratio under a factorised or non-factorised approximate posterior.

\subsection{Exact marginalisation over discrete latent variables}

Notably, the TMC framework can be extended to incorporate exact marginalisation over discrete variables.
In particular, we take the number of importance samples, $K_i$, to be equal to the number of settings of the discrete variable, we consider a uniform proposal, $\Q{z_i}=1/K_i$, and we use stratified sampling, such that each possible setting of the latent variable is represented by one sample (e.g.\ taking $z_i = \{1,2,\dotsc,K_i\}$, we might have $z_i^\k{i} = k_i$).
Making these choices, and taking $z_1$ to be a discrete variable, the TMC estimator over $\P{x, z_1^\k{1}, z_2^\k{2},\dotsc,z_n^\k{n}}$ is,
\begin{align*}
  \mathcal{P}_\text{TMC} &= \frac{1}{K^{n-1}} \sum_{k_2,\dotsc,k_n} \sum_{z_1} \frac{\P{x, z_1, z_2^\k{2},z_3^\k{3}\dotsc,z_n^\k{n}}}{\Q{z_2^\k{2}}\Q{z_3^\k{3}}\dotsm\Q{z_n^\k{n}}}.
\end{align*}
Note that this is exactly equal to a different TMC estimator, over a model with $z_1$ marginalised out (i.e.\ $\P{x, z_2, z_3, \dotsc, z_n}$).
This is important because it enables us to link TMC with the rich prior literature on exact marginalisation in discrete graphical models, and because it allows us to combine importance sampling over continuous variables and exact marginalisation over discrete variables into a single framework.

\subsection{Numerically stable matrix (tensor) products in the log-domain}

When we take inner products of tensors representing large probabilities, there is a considerable risk of numerical overflow.
To avoid this risk, we work in the log-domain, and write down a numerically stable matrix-inner product, denoted logmmexp, by analogy with the standard logsumexp function.
In particular, consider the problem of computing $e^{Z_{ik}}$, as the matrix product of $e^{X_{ij}}$ and $e^{Y_{jk}}$,
\begin{align*}
  e^{Z_{ik}} &= \sum_j e^{X_{ij}} e^{Y_{jk}}.
  \intertext{taking the logarithm so as to compute $Z_{ik}$,}
  Z_{ik} &= \log \b{\sum_j e^{X_{ij}} e^{Y_{jk}}}.
  \intertext{as the elements of $X_{ij}$ and $Y_{jk}$ could be very large (or very small), to ensure numerical stability of the sum, we add and subtract $x_i$ and $y_k$,}
  Z_{ik} &= \log \b{\sum_j e^{X_{ij} - x_i} e^{Y_{jk} - y_k}} + x_i + y_k,
\end{align*}
  where
\begin{align*}
  x_i &= \max_j X_{ij}\\ 
  y_k &= \max_j Y_{jk}.
\end{align*}

\subsection{Combining DReGs and TMC}
\label{sec:dregs}
To perform DReGs, we optimize the generative parameters using the usual IWAE/TMC cost function, but use a different strategy for optimizing the recognition model, that involves non-trivial manipulations of the importance weights.
In particular, the DReGs recognition updates are given by,
\begin{align}
  \label{eq:dregs}
  \sum_i & \frac{w_i^2}{(\sum_j w_j)^2} \d[z_i]{\phi} \d[\log w_i]{z_i} \\
  \nonumber
  &= \sum_i \frac{w_i^2}{(\sum_j w_j)^2} \d[z(\epsilon_i; \phi)]{\phi} \left. \d[\log w(z; \phi)]{z} \right\rvert_{z=z(\epsilon_i; \phi)},
\end{align}
where the first version is that given in prior work, and the second version has been written out more carefully to highlight the functional dependencies, and hence how the partial derivative applies to each term.
The latents have been written in their usual reparameterised form, 
and the importance weights can be written as a function of the latents, and the parameters,
\begin{align}
  w(z; \phi) &= \frac{\P{x, z}}{\Q[\phi]{x, z}},
  \intertext{but we write down the individual importance weights as functions of the reparameterised noise, $\epsilon_i$, and the parameters,}
  \label{eq:dregs:wi}
  w_i(\epsilon_i; \phi) &= w(z(\epsilon_i; \phi); \phi).
\end{align}
We cannot compute this function directly in the TMC set-up, because TMC involves exponential numbers of importance samples, and allows only a fairly restricted set of operations (such as summing the importance weights) to be performed efficiently. 
In contrast, DReGs appears to require complex, almost arbitrary operations over the importance weights.
However, it is possible to write down a surrogate objective, utilizing the stop-gradients operation, that does have the required gradients.
To do so, we need to begin by carefully introducing notation.
Remember that any particular importance weight, $w_i$, can be written as a function of the uniform noise, $\epsilon_i$ and the parameters (Eq.~\eqref{eq:dregs:wi}), and
as such, the gradient of $w_i$ can be broken up into two terms, 
\begin{align}
  \label{eq:dwi}
  \d[w_i(\epsilon_i; \phi)]{\phi} &= \d[w(z_i; \phi)]{\phi}  + \d[z(\epsilon_i; \phi)]{\phi} \left. \d[w(z; \phi)]{z} \right|_{z=z(\epsilon_i; \phi)}
\end{align}
where the first term is the direct effect of $\phi$, on $w_i$, and the second term is the ``indirect'' effect, through the reparameterised latents.
Now we are in a position to define $\wh_i$ and $\wb_i$, which always have the same value as $w_i$, but where we have applied the stop-gradients operation to drop different gradient terms.
In particular, $\wb_i$ stops all gradients, so that (in a slight abuse of notation),
\begin{align}
  \d[\wb_i]{\phi} &= 0,
\end{align}
and $\wh_i$ stops only the ``direct'' term in Eq.~\eqref{eq:dwi}, such that,
\begin{align}
  \d[\wh_i]{\phi} &= \d[z(\epsilon_i; \phi)]{\phi} \left. \d[w(z; \phi)]{z} \right|_{z=z(\epsilon_i; \phi)}
\end{align}
Now, we hypothesise that the DReGs estimator can be rewritten as,
\begin{align}
  \frac{1}{2} \b{\frac{\sum_j \wb_j^2}{\b{\sum \wb_j}^2}} \log \sum_i \wh_i^2
\end{align}
Note that this is written entirely in terms of $\sum_j w_j$, which can be computed directly using TMC, as given above, and $\sum_j w_j^2$, which can be computed by running TMC again, with squared weights.
The gradient of $\wb_j$ is zero, so
\begin{align}
  \d{\phi} & \sb{\frac{1}{2} \b{\frac{\sum_j \wb_j^2}{\b{\sum \wb_j}^2}} \log \sum_i \wh_i^2}\\
  &= \frac{1}{2} \b{\frac{\sum_j w_j^2}{\b{\sum w_j}^2}} \d{\phi} \log \sum_i \wh_i^2,
  \intertext{where the value of $w_j$, $\wh_j$ and $\wb_j$ is equal, so when the gradient operation is no longer applied, we can revert to the standard notation, $w_j$.  Applying the derivative to the logarithm,}
  &= \frac{1}{2} \b{\frac{\sum_j w_j^2}{\b{\sum w_j}^2}} \frac{\sum_i \d{\phi} \wh_i^2}{\sum_j w_j^2},\\
  \intertext{cancelling terms,}
  &= \frac{1}{2} \frac{\sum_i \d{\phi} \wh_i^2}{\b{\sum w_j}^2},\\
  \intertext{and applying the derivative to $\wh_j^2$,}
  &= \frac{\sum_i w_i \d{\phi} \wh_i}{\b{\sum w_j}^2},\\
  \intertext{Finally, using $\d{\phi} \wh_i = w_i \d{\phi} \log \wh_i$,}
  &= \sum_i \frac{w_i^2}{\b{\sum w_j}^2} \d{\phi} \log \wh_i,
  \intertext{and substituting the gradient of $\wh_i$,}
  &= \sum_i \frac{w_i^2}{\b{\sum w_j}^2} \d[z_i]{\phi} \d[\log w_i]{z_i},
\end{align}
which matches Eq.~\eqref{eq:dregs}, as required.

\end{document}